%% file: main.tex
\documentclass[conference]{IEEEtran}
\IEEEoverridecommandlockouts
\usepackage{cite}
\usepackage{amsmath,amssymb,amsfonts}
\usepackage{algorithmic}
\usepackage{graphicx}
\usepackage{textcomp}
\usepackage{xcolor}

\usepackage{listings}
\usepackage{booktabs}
\usepackage{multirow}
\usepackage{subfigure}
\usepackage{caption}

\newcommand{\linebreakand}{%
  \end{@IEEEauthorhalign}
  \hfill\mbox{}\par
  \mbox{}\hfill\begin{@IEEEauthorhalign}
}

\def\BibTeX{{\rm B\kern-.05em{\sc i\kern-.025em b}\kern-.08em
    T\kern-.1667em\lower.7ex\hbox{E}\kern-.125emX}}
\begin{document}

\title{Billion-scale Pre-trained E-commerce Product Knowledge Graph Model}

\author{
\IEEEauthorblockN{ Wen Zhang\textsuperscript{*}}
\IEEEauthorblockA{
\textit{Zhejiang University}\\
Hangzhou, China \\
wenzhang2015@zju.edu.cn\\
}
\and
\IEEEauthorblockN{Chi-Man Wong\textsuperscript{*}}
\IEEEauthorblockA{
\textit{Alibaba Group,} \\
\textit{University of Macao} \\
chiman.wcm@alibaba-inc.com \\
}
\and
\IEEEauthorblockN{ Ganqiang Ye}
\IEEEauthorblockA{
\textit{Zhejiang University}\\
Hangzhou, China \\
yeganqiang@zju.edu.cn\\
}
\and 
\IEEEauthorblockN{ Bo Wen}
\IEEEauthorblockA{
\textit{Zhejiang University}\\
Hangzhou, China \\
wenbo1@zju.edu.cn \\
}
\and
% \linebreakand
\IEEEauthorblockN{ Wei Zhang}
\IEEEauthorblockA{
\textit{Alibaba Group}\\
Hangzhou, China \\
lantu.zw@alibaba-inc.com \\
}
\and 
\IEEEauthorblockN{ Huajun Chen\textsuperscript{\textsection}}
\IEEEauthorblockA{
\textit{College of Computer Science}\\
\textit{Hangzhou Innovation Center}\\
\textit{Zhejiang University} \\
\textit{AZFT Joint Lab for Knowledge Engine} \\
huajunsir@zju.edu.cn \\
}
}
\maketitle
\begingroup\renewcommand\thefootnote{*}
\footnotetext{Equal contribution.}
\begingroup\renewcommand\thefootnote{\textsection}
\footnotetext{Corresponding author.}

\begin{abstract}

In recent years, knowledge graphs have been widely applied to organize data in a uniform way and enhance many tasks that require knowledge, for example, online shopping which has greatly facilitated people's life. As a backbone for online shopping platforms, we built a billion-scale e-commerce product knowledge graph for various item knowledge services such as item recommendation. However, such knowledge services usually include tedious data selection and model design for knowledge infusion, which might bring inappropriate results. Thus, to avoid this problem, we propose a Pre-trained Knowledge Graph Model (PKGM) for our billion-scale e-commerce product knowledge graph, providing item knowledge services in a uniform way for embedding-based models without accessing triple data in the knowledge graph. Notably, PKGM could also complete knowledge graphs during servicing, thereby overcoming the common incompleteness issue in knowledge graphs. We test PKGM in three knowledge-related tasks including item classification, same item identification, and recommendation. Experimental results show PKGM successfully improves the performance of each task.

\end{abstract}


\input{C1-Introduction}
\input{C2-Method}
\input{C3-Experiment}
\input{C4-Related_work}

\input{C5-Conclusion}
\input{C6-Acknowledgements}

\bibliographystyle{IEEEtran}
\bibliography{main}

\end{document}

%% file: C1-Introduction.tex
\section{Introduction}
Online shopping has greatly contributed to the convenience of people's lives. More and more retailers open online shops to attract increasing people favoring online-shopping, thus the e-commerce era witnessed rapid development in the past few years. The growing transaction on e-commerce platforms is based on billions of items of all kinds, which should be well organized to support the daily business. 

Knowledge Graphs (KGs) represent facts as triples, such as \emph{(iPhone, brandIs, Apple)}. Due to the convenience of fusing data from various sources and building semantic connections between entities, KGs have been widely applied in industry and many huge KGs have been built (e.g., Google's Knowledge Graph, Facebook's Social Graph, and Alibaba/Amazon's e-commerce Knowledge Graph). We built an e-commerce product knowledge graph and make it a uniform way to integrate massive information about items on our platform, which helps us build an integrated view of data. Notably, the e-commerce product knowledge graph contributes to a variety of tasks ranging from searching, question answering, recommendation, and business intelligence, etc. Currently, our product KG contains 70+ billion triples and 3+ million rules. 

Our product KG has been widely used for knowledge-enhanced e-commerce applications, such as item attributes prediction, same item identification, item category prediction, and item recommendation. These are key tasks for managing more complete and clean data and providing better item services for customers. In supervised paradigms, these knowledge-enhanced tasks rely heavily on data that are vital for them, for example, item title for same item identification and user-item interaction bipartite graph for recommendation. Apart from these data, information contained in the product knowledge graph about items‘ attributes such as brands, series, and colors, are valuable side information to enhance these tasks. 

To serve these knowledge-enhanced tasks, we used to provide item information formed as triples in product KG to facilitate knowledge enhancement. While it not only results in tedious data selection working for us, mainly refers to neighbor sampling, but also leaves tedious knowledge enhancement model design for downstream use. More importantly, similar to other KGs, our product KG is still far from complete and the key missing information may bias those downstream tasks. 

Recently, a variety of knowledge graph embedding methods\cite{TransE-Bordes-2013} are proposed for completion, which learn representation for entities and relations in continuous vector spaces, called embeddings, and infer the truth value of a triple via embedding calculation. Embeddings usually encode global information of entities and relations and are widely used as general vector space representations for different tasks. Thus for services of knowledge-enhanced tasks, instead of providing triple data, we are wondering that \textbf{is it possible to make embeddings from product KG as a uniform knowledge provider for knowledge-enhanced tasks, which not only avoid tedious data selection and model design, but also overcome the incompleteness of product KG.}

Since the concept of `pre-training and fine-tuning' has proven to be very useful in the Natural Language Processing community \cite{BERT}, referring to pre-training a language model with a huge amount of language data and then fine-tune it on downstream tasks such as sentiment analysis \cite{ACL2020_Pretrain4SA}, relation extraction \cite{ACL2019_Pretrain4RE}, text classification \cite{EMNLP2019_Pretrain4TC} and so on. We also consider to pre-train our product KG and make it possible to conveniently and effectively serve downstream tasks requiring background knowledge about items. The convenience refers that downstream tasks do not need to revise their model much to adapt to background knowledge. The effectiveness refers that downstream tasks enhanced with knowledge could achieve better performance, especially with a small amount of data. 

The purpose of \textbf{P}re-trained \textbf{K}nowledge \textbf{G}raph \textbf{M}odel (\textbf{PKGM}) is to provide knowledge services in continuous vector space, through which downstream tasks could get necessary fact knowledge via embedding calculation without accessing triple data. The key information of product KG concerned by downstream tasks mainly includes two parts: (1) whether one item has a given relation(attribute), (2) what is the tail entity(property value) of given an item and relation(attribute). Thus making the incompleteness issue of knowledge graph also into consideration, a pre-trained knowledge graph model should be capable of: (1) showing whether an entity has a certain relation, (2) showing what is the tail entity for a given entity and relation, and (3) completing the missing tail entity for a given entity and relation if it should exist.

In this paper, we introduce a pre-trained e-commerce product knowledge graph model with two modules. One is Triple Query Module to encode the truth value of an input triple, which could serve for tail entity queries after training. Since a lot of knowledge graph embedding methods are proposed for triple encoding and link prediction, we apply the simple and effective TransE \cite{TransE-Bordes-2013} in the triple query module. It represents entities and relations as vectors and restricts the addition of head entity embedding and relation embedding should approach tail entity embedding if the input triple is true. The other one is Relation Query Module to encode the existence of a relation for an entity, which could serve for queries of relation existence after training. In this module, we take entity and relation embeddings from the triple query module and create an entity transformation matrix $\mathbf{M}_r$ for each relation $r$. With $\mathbf{M}_r$, we make the transformed head entity embedding $\mathbf{h}^\prime$ approaches to relation embedding $\mathbf{r}$ if $h$ owns relation $r$. 

After pre-training, the triple query module and the relation query module could provide knowledge service vectors for any target entity. Triple query module provides service vectors containing tail entity information of target entity and relation. For example, suppose the target entity as a smartphone, the triple query module will provide predicted tail entity embedding with target relations such as \emph{brandIs, seriesIs, memoryIs}, etc. , no matter if these triples exist in the knowledge graph or not. Thus advantages of the triple query module are that (1) it could provide knowledge service based on computation in vector space without accessing triples data, (2) it could complete the tail entity of target entity and relation during service. Relation query module provides service vectors containing existence information of different relations for target entities, in which the corresponding service vector will approach zero vector if the target entity (should) have the relation. Combining service vectors from triple query module and relation query module, information could be provided without executing symbolic queries by the pre-trained knowledge graph model are more than triples contained in the product knowledge graph. 
 
With service vectors, we propose general ways to incorporate them in embedding-based methods, called base models, for downstream tasks. We classify embedding-based methods into two types according to the number of input embeddings of target entities. The first type refers to methods with \emph{a sequence of embeddings} for target entities, and the second one refers to \emph{a single embedding} for target entities. We propose a general way to apply service vectors into each of the two types of methods. For sequence-embedding inputs, we extend the input sequence by appending service vectors after original inputs of target entities. And for single-embedding inputs, we condense service vectors into one and concatenate it with the original input for target entities. 

We pre-train the billion-scale product knowledge graph with PKGM model and apply service vectors to enhance three item related tasks, item classification, same product identification and item recommendation, following our proposed servicing and application methods. The experimental results show that PKGM successfully improves performance of these three tasks with a small amount of training data. 

In summary, contributions of this work are as follows: 
\begin{itemize}
    \item We propose a way to pre-train a knowledge graph, which could provide knowledge services in vector space to enhance knowledge-enhance downstream tasks in a general way. 
    \item Our proposed PKGM has two significant advantages, one is completion capability, and the other one is triple data independency.
    \item We practice PKGM on our billion-scale product knowledge graph and test it on three tasks, showing that PKGM successfully enhances them with item knowledge and improves their performance.
\end{itemize}

%% file: C2-Method.tex
\section{Pre-trained Knowledge Graph Model (PKGM)}

Product knowledge graph $\mathcal{K} = \{\mathcal{E}, \mathcal{R}, \mathcal{F}\}$ stores information of items and relations between items as triples, in which $\mathcal{E}, \mathcal{R}$ and $\mathcal{F}$ are entity set, relation set and triple set respectively. $\mathcal{E} = \{\mathcal{I, V}\}$ contains a set of items $\mathcal{I}$ and a set of values $\mathcal{V}$. $\mathcal{R} = \{ \mathcal{P}, \mathcal{R^\prime} \}$ contains a set of items properties $\mathcal{P}$ and a set of relations between items $\mathcal{R}^\prime$. Our product knowledge graph contains more than $70$ billion triples. 

\begin{figure*}
    \centering
    \includegraphics[width = \textwidth]{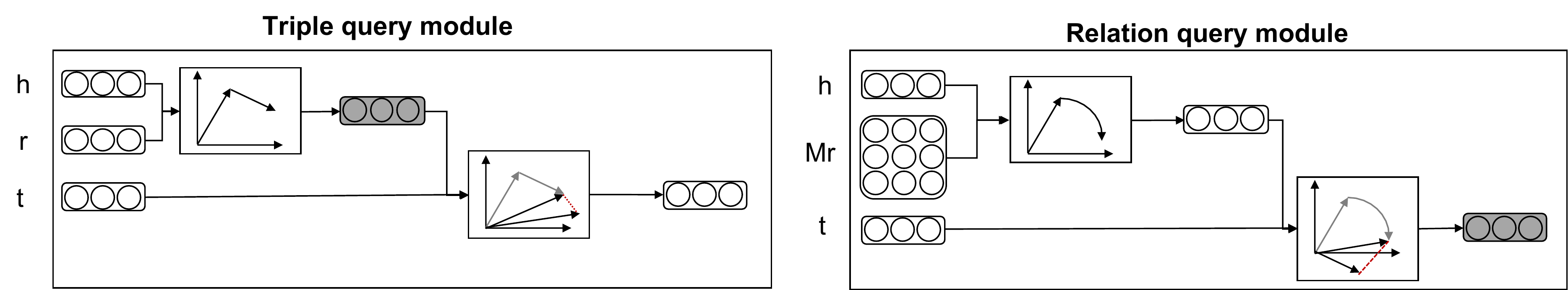}
    \caption{Pre-trained Knowledge Graph Model.}
    \label{fig:PKGM}
\end{figure*}

We propose a pre-trained knowledge graph model to encode $\mathcal{K}$ into continuous vector space, making it possible to provide knowledge services for knowledge-enhanced downstream tasks based on calculation in vector space. Knowledge services for other tasks usually refer to returning data in knowledge graph matching knowledge queries. Triple query and relation query are two types of basic queries commonly executed in knowledge graphs. Triple queries are in the following form
\begin{lstlisting}[frame=shadowbox]
SELECT ?t
WHERE {h r ?t}
\end{lstlisting}
where the target is to get the tail entity $t$ given an head entity $h$ and relation $r$. Relation queries are
\begin{lstlisting}[frame=shadowbox]
SELECT ?r
WHERE {h ?r ?t}
\end{lstlisting}
where the target is to get relations of a given entity $h$. 
With entity and relation sets known, combining these two types of queries, we could recover all triples in a knowledge graph.

In PKGM, we build two query modules to simulate information queries of knowledge graphs in continuous vector space. One is Triple Query Module for triple queries, which maintains an assumption to transfer the head entity embedding and relation embedding to tail entity embedding in vector space, expressed as $f_{triple}(\textbf{h}, \textbf{r}) \approx \textbf{t}$. The other one is Relation Query Module for relation queries. It maintains an assumption to show whether an entity should have a relation or not. In vector space, we make zero vector $\mathbf{0}$ represent EXISTING, thus the function in Relation Query Module should be  $f_{rel}(\mathbf{h}, \mathbf{r}) \approx \mathbf{0}$ if exists. 

\subsection{Triple query module}
The purpose of the triple query module is to transfer the head entity and  relation to the tail entity and encode the truth value of a triple. Since a lot of knowledge graph embedding methods have been proposed for this task, we apply TransE \cite{TransE-Bordes-2013} for its simplicity and effectiveness. 

Each $e \in \mathcal{E}$ and $r \in \mathcal{R}$ is embedded as a vector.  Following the translation assumption, for each positive triple $(h, r, t)$, we make $\mathbf{h} + \mathbf{r} \approx \mathbf{t}$, where $\mathbf{h}\in \mathbb{R}^d, \mathbf{r}\in \mathbb{R}^d$  and $\mathbf{t}\in \mathbb{R}^d$ are embedding of $h, r$ and $t$ respectively. Score function for $(h, r, t)$ is 
\begin{equation}
f_{triple}(h,r,t) = \|\mathbf{h} + \mathbf{r} - \mathbf{t}\|  
\label{score-triple}
\end{equation}
where $\| \mathbf{x} \|$ is the L1 norm of  $\mathbf{x}$. It should make $\mathbf{h} + \mathbf{r}$ approach $\mathbf{t}$ if $(h,r,t)$ is  positive and  far away otherwise.

\subsection{Relation query module}
The purpose of relation query module is to encode whether an entity should have one relation or not. We represent the concept of  EXIST as $\mathbf{0}$ in vector space.  To encode the existence of relation $r$ for $h$, we design a function $f_{rel}(\mathbf{h}, \mathbf{t})$ and make $f_{rel}(\mathbf{h}, \mathbf{t}) \approx \mathbf{0}$ if $h$ (should) have  $r$ and keep $f_{rel}(\mathbf{h}, \mathbf{t})$ far away from $\mathbf{0}$ otherwise. 

We devise a transferring matrix for each relation $r \in \mathcal{R}$, denoted as $\mathbf{M}_r$. With $\mathbf{M}_r$ we could transfer $\mathbf{h}$ to $\mathbf{r}$. More specifically, the discrepancy between the transferred head entity representation and relation embedding should be close to $\mathbf{0}$ if $h$ has relation $r$ or should have. And the discrepancy should be far away from $\mathbf{0}$ if $h$ does not have or should not have relation $r$. Thus $f_{rel}$ is designed as follows
\begin{equation}
f_{rel}(h, r) = \| \mathbf{M}_r \mathbf{h} - \mathbf{r}\|
\label{score-rel}
\end{equation}
For a positive pair $(h,r)$, $f_{rel}(h,r)$ should be as small as possible, while be as large as possible for negative ones. 

\subsection{Loss function}
The final score of the given triple $(h,r, t)$ is 
\begin{equation}
    f(h,r,t) = f_{triple}(h,r,t) + f_{rel}(h,r)
\end{equation}
in which positive pairs should have a small score and negative pairs should have a large one. A margin-based loss is set as a training objective for PKGM
\begin{equation}
L = \sum_{(h,r,t) \in \mathcal{K}}  [f(h,r,t) + \gamma - f(h',r',t') ]_+
\label{loss}
\end{equation}
where $(h',r',t')$ is a negative triple generated for $(h,r,t)$ by randomly sample an entity $e \in \mathcal{E}$ to replace $h$ or $t$ , or randomly sample a relation $r' \in \mathcal{R}$ to replace $r$. $\gamma$ is a hyperparameter representing the margin which should be achieved between  positive sample scores and negative sample scores. And 
\begin{equation}
[x, y]_+=\left\{
\begin{aligned}
x &  \text{\; if \; } x \ge y \\
y & \text{\; if \; } x < y 
\end{aligned}
\right.
\end{equation}

\subsection{Service}
After training, PKGM has (1) good model parameters including entity embeddings, relation embeddings and transferring matrices, (2) functions in  triple query module and relation query module. Then it could provide service for triple queries and relation queries in vector space. 

\subsubsection{Service for triple queries}
PKGM provides the representation of candidate tail entity, when given a head entity $h$ and a relation $r$ as follows:
\begin{equation}
    S_{triple}(h,r) = \mathbf{h} + \mathbf{r} 
\end{equation}
with which $S_{triple}(h,r)$ will approximate to $\mathbf{t}$ if $(h,r,t) \in \mathcal{K}$ as a result of training objective (Equation(\ref{loss})), and the output of $S_{triple}(h,r)$ will be an entity representation that is most likely to be the right tail entity if there is no triple existing with $h$ as head entity and $r$ as relation in $\mathcal{K}$. This is the inherent completion capability of knowledge graph embedding methods, which has been widely proved and applied for knowledge graph completion\cite{DBLP:journals/corr/abs-2002-00388}. We summarize the function for pre-training and servicing in Table~\ref{tab:PKGM-functs}.

\input{tables/PKGM}

\subsubsection{Service for relation queries}
PKGM provides a vector representing whether an entity $h$ has or should have triples about relation $r$ as follows:
\begin{equation}
    S_{rel}(h, r) = \mathbf{M}_r \mathbf{h} - \mathbf{r} 
\end{equation}
There are three situations: (1) if $h$ links to entities according to $r$, then  $S_{rel}(h, r)$ will approximate to EXIST embedding $\mathbf{0}$, (2) if $h$ does not link to any entity according to $r$ but it should, then $S_{rel}(h, r)$ will also approximate to EXIST embedding $\mathbf{0}$, (3) if $h$ does not link to any entity according to $r$ and it should not, $S_{rel}(h, r)$ will be away from  $\mathbf{0}$. This is the relation completion capability of PKGM gaining from Equation (\ref{score-rel}). 

There are two significant advantages  of getting knowledge via PKGM's query services given $h$ and $r$: 
\begin{itemize}
\item We could access the tail entity in an implicit way via calculation in vector space without truly querying triples existing in the knowledge graph. This makes query service independent with data and ensures data protection. 
\item Results for each input pair are uniformed as two vectors from two query modules instead of triple data,  making it easier to apply them in downstream use models since designing models to encode triple data is avoided.  
\item We could get the inferred tail entity $t$ even triple $(h,r,t) \notin \mathcal{K}$, which greatly overcomes the incompleteness disadvantages of KG.   
\end{itemize}

\subsection{Applying service vectors in downstream use models}
With service vectors from PKGM, we introduce two general ways to apply them in embedding-based models for downstream use. We categorize embedding-based models into two classes according to the number of input embeddings for an entity in the model, one is inputting a sequence of  embeddings and the other one is inputting a single embedding. Given an item as target entity in the context of e-commerce, service vectors from triple query modules are denoted by $[S_1, S_2, ..., S_k]$ and those from relation query module denoted by $[S_{k+1}, S_{k+2}, ..., S_{2k}]$. 

\begin{figure}[!hbpt]
    \centering
    \includegraphics[width=0.49\textwidth]{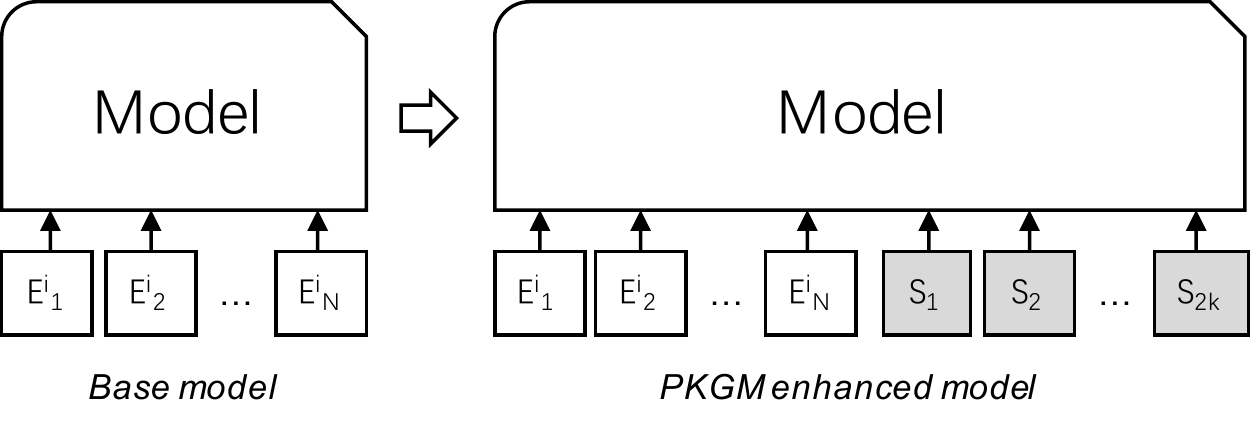}
    \caption{Application of PKGM for models with sequence embedding inputs.}
    \label{fig:PKGM+1}
\end{figure}

The idea of integrating service vectors from PKGM into sequence models is shown in Fig.\ref{fig:PKGM+1}. 
In these models, the sequential embedding for one $item_i$ are usually generated according to item side information, like word embedding of description text or labeled features. And these models are capable of automatically capturing interactions between embeddings in the sequence, thus we propose to append all service vectors for $item_i$ from PKGM at the end of the input sequence. The model will make service vectors interact with other input embeddings automatically. Supposing that the original input is $[E_1^i, E_2^i, ..., E_N^i]$, after appending service vectors, the input will be  $[E_1^i, E_2^i, ..., E_N^i, S_1, S_2, ..., S_{2k}]$, where we first append service vectors from triple query module and then append those from relation query module.

\begin{figure}[!htbp]
    \centering
    \includegraphics[width=0.49\textwidth]{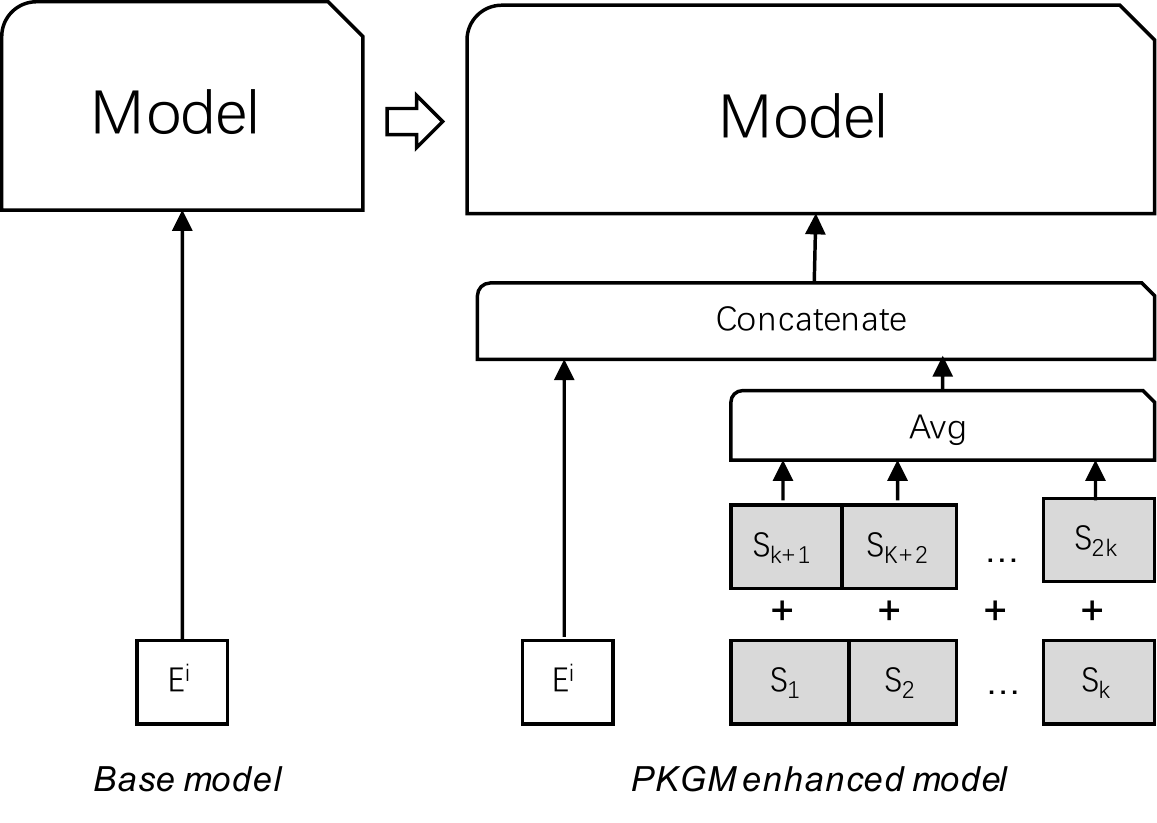}
    \caption{Application of PKGM for models with single embedding input.}
    \label{fig:PKGM+2}
\end{figure}

The idea of integrating service vectors from PKGM into  models with single embedding input for an entity is shown in Fig.\ref{fig:PKGM+2}. In these models, the single embedding usually refers to the embedding of $item_i$ in current latent vector space and is learnt during training. To keep the informational balance for an entity between embeddings from the original model and our PKGM, we propose to firstly integrate all service vectors for $item_i$ from PKGM into one. During the integration, service vectors corresponding to the same relation from two modules should be considered together, thus we firstly make a concatenation with service vectors from triple query module and relation query module as follows:
\begin{equation}
    S_j^\prime = [S_j;S_{j + k}], \text{where}\; j \in [1, k])
\end{equation}
where $[x;y]$ means concatenating vector $x$ and $y$. Then we integrate them into one as follows:
\begin{equation}
    S = \frac{1}{k}\sum_{j \in [1, k]} S_j
\end{equation}

Then we concatenate $S$ with original item embedding $E_i$ as the single input. 

%% file: tables/PKGM.tex
\begin{table}[t]
    \centering
    \caption{Functions for pre-training and servicing stage of PKGM. The subscript letter $T$ and $R$ denotes $triple$ and $rel$.}
    \begin{tabular}{c|c | c}
    \toprule
        Module & Pre-training & Servicing \\
         \midrule
        Triple 
        & $f_{T}(h,r,t) = \|\mathbf{h} + \mathbf{r} - \mathbf{t}\|$    
        & $S_{T}(h,r) = \mathbf{h} + \mathbf{r}$ \\
        \midrule
        % \multirow{2}{*}{Relation}
        % & \multirow{2}{*}{$f_{R}(h, r) = \| \mathbf{M}_r \mathbf{h} - \mathbf{r}\| $} 
        % & $S_{R}(h, r) = \mathbf{M}_r \mathbf{h} - \mathbf{r}$\\
        % \cline{3-3}
        % & &$S_{R}(h, r) = \| \mathbf{M}_r \mathbf{h} - \mathbf{r}\| $  \\
        Relation
        & $f_{R}(h, r) = \| \mathbf{M}_r \mathbf{h} - \mathbf{r}\| $ 
        & $S_{R}(h, r) = \mathbf{M}_r \mathbf{h} - \mathbf{r}$\\
        % \cline{3-3}
        % & &$S_{R}(h, r) = \| \mathbf{M}_r \mathbf{h} - \mathbf{r}\| $  \\
        \bottomrule
    \end{tabular}
    \label{tab:PKGM-functs}
\end{table}

%% file: C3-Experiment.tex
\section{Experiments}

\input{C3-E-1pre-training}
\input{C3-E-2Item_classification}
\input{C3-E-3Same_product_identification}

\input{C3-E-4Item_recommendation}

%% file: C3-E-1pre-training.tex
\subsection{Pre-training}

\subsubsection{Dataset}
We pre-train PKGM on a sub dataset of our billion-scale E-commerce product knowledge graph (PKG) called PKG-sub. PKG has a total number of 1.2B attribute records for 0.2B items, where all the attributes are filled by sellers in Taobao. We process the data into triplet format using Alibaba Maxcompute, a max-reduce framework. We remove the attributes with occurrences less than 5000 in PKG, since these attributes are likely noisy that not only increase largely the model size but also deteriorate the model performance. The statistic details of PKG-sub are shown in Table~\ref{tab:dataset-pre-train}.   
\input{tables/kg-dataset}

For each item $item_i$ in the dataset, we select $10$ key relations for it according to its category. More specifically, suppose $item_i$ belongs to category $C$, we gather all items belonging to $C$ and account for the frequency of properties in those items, then select top $10$ most frequent properties as key relations. After pre-training, our PKGM will provide service vectors for target items with regard to those key relations. 

\subsubsection{Training details}
Our PKGM is implemented with Tensorflow\cite{tensorflow} and Graph-learn\footnote{https://github.com/alibaba/graph-learn}. Graph-learn~\cite{zhu2019aligraph} is a large-scale distributed framework for node and edge sampling in graph neural networks. We use Graph-learn to perform edge sampling, and 1 negative sample is generated for each edge.  For model setting, Adam\cite{adam} optimizer is employed with initial learning rate as 0.0001, training batch size as 1000, and embedding dimension as 64. Finally the model size is 88GB, and we train it with 50 parameter servers and 200 workers for 2 epochs. The whole training process consumed 15 hours.

%% file: tables/kg-dataset.tex
\begin{table}[!hbpt]
    \centering
    \caption{Statistics of PKG-sub for pre-training.}
    \begin{tabular}{ c | c |c|c | c}
    \toprule
           & \# items & \# entity & \# relation & \# Triples \\
           \midrule 
           PKG-sub &142,634,045 &142,641,094 &426 &1,366,109,966
 \\
        \bottomrule
    \end{tabular}
    \label{tab:dataset-pre-train}
\end{table}

%% file: C3-E-2Item_classification.tex
\subsection{Item Classification}
\subsubsection{Task definition}
The target of item classification is to assign an item to a class in the given class list. Item classification is a common task in our e-commerce platform since new tags and features are added frequently to items, such as lifestyle and crowd that items are suitable for. For different classification tasks, the candidate class varies.  For example, candidate classes for lifestyle classification include Vacation, Yoga, Hiking, and so on, and those for crowd classification include Programmer, Parents, Girlfriend, and so on. 

Usually, item titles are used for classification since most items in our platform have a title filled by managers of online shops. Thus we could frame it as a text classification task. Under supervised training, the target is to train a mapping function $f: \mathcal{T} \mapsto \mathcal{C}$ where $\mathcal{T}$ is a set of item titles and $\mathcal{C}$ is a set of classes. For each title  $t \in \mathcal{T}$, $t = [ w_1, w_2, w_3, ..., w_n ]$ is an ordered sequence composed of words. 

\subsubsection{Model}

\textbf{Base model}
Text classification is an important task in Natural Language Processing (NLP) and text mining, and many methods have been proposed for it. In recent years, the deep learning model has been shown to outperform traditional classification methods\cite{DBLP:conf/acl/HenaoLCSSWWMZ18, BERT}. Given the input document, the mapping function $f$ firstly learns a dense representation of it via representation learning and then uses this representation to perform final classification. Recently, large-scale pre-trained language models, such as ELMo\cite{ELMo}, GPT\cite{GPT} and BERT\cite{BERT}, have become the \emph{de facto} first representation learning step for many natural language processing tasks. Thus we apply BERT as the base model for item classification. 

BERT is a pre-trained language model with multi-layers of bidirectional encoder representations from Transformers, trained with unlabeled text by jointly conditioning on both left and right context in all layers. BERT is trained on a huge amount of texts with masked language model objective. Since the use of BERT has become common and we conduct item classification experiments with released pre-trained BERT by Google. We will omit an exhaustive background description of the model architecture and refer the readers to \cite{BERT} as well as the excellent guides and codes\footnote{https://github.com/google-research/bert} of fine-tuning BERT on downstream tasks. 

\begin{figure*}[!htbp]
     \centering
     \subfigure[BERT.]{
         \centering
         \includegraphics[width=0.435\textwidth]{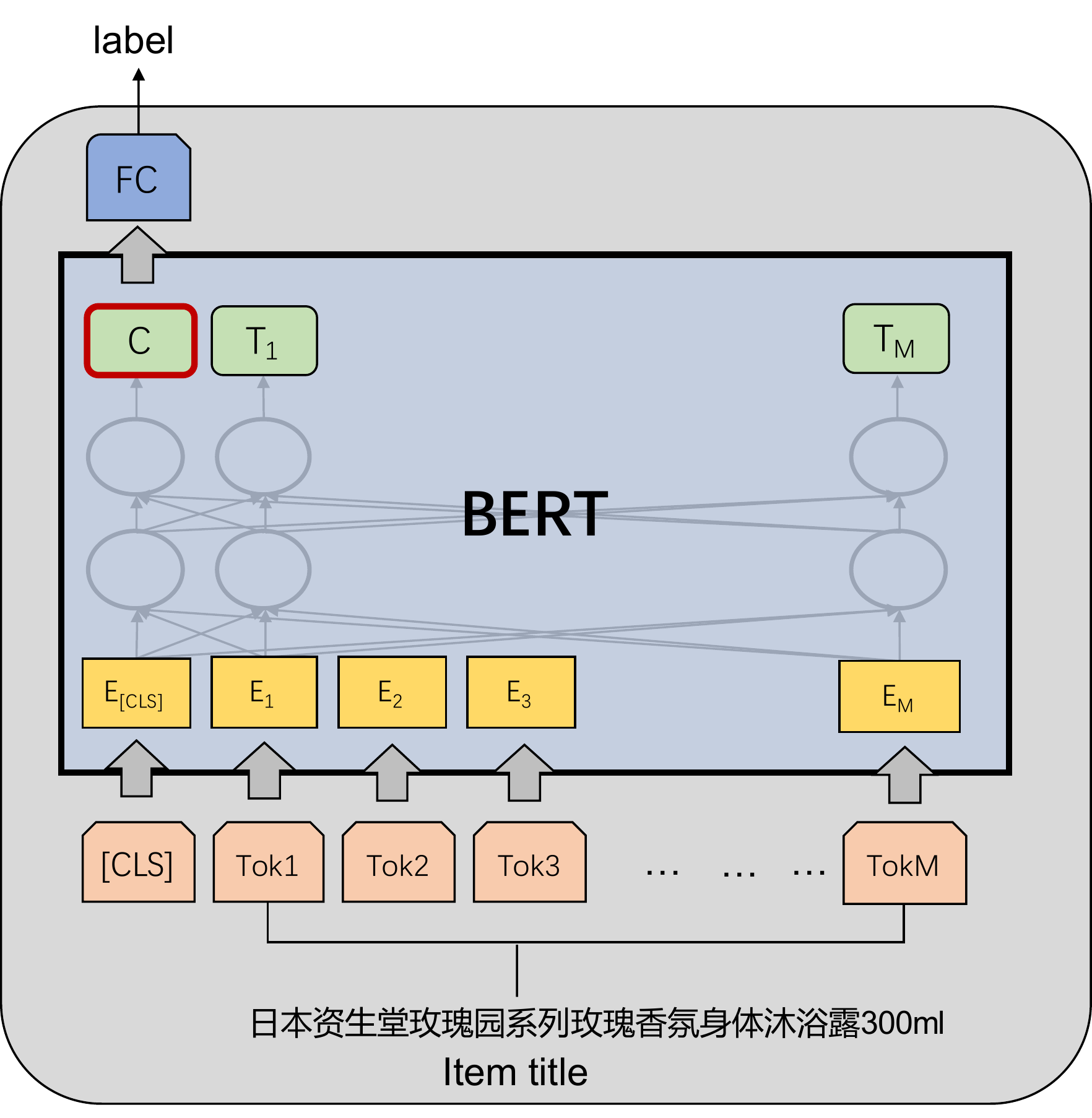}
         \label{fig:cate-base}
     }
     \subfigure[BERT$_{PKGM}$.]{
         \centering
         \includegraphics[width=0.45\textwidth]{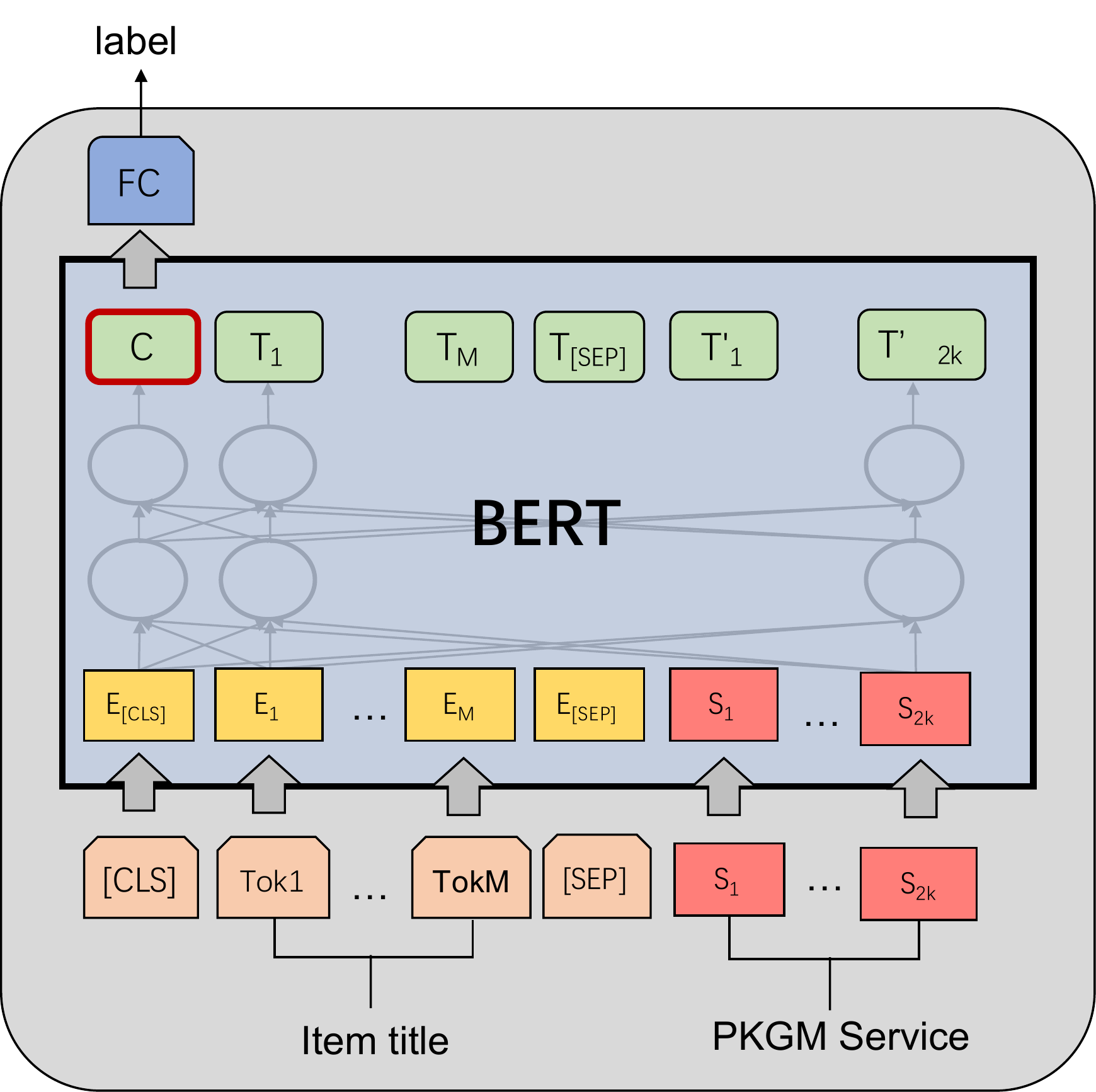}
         \label{fig:cate-base+PKGM}
     }
    \caption{Models for item classification.}
    \label{fig:model-classification}
\end{figure*}

We show details of applying BERT on item classification task in Figure~\ref{fig:model-classification}. We input the text sequence of title into BERT and take out the representation corresponds to [CLS] symbol $C$. To predict the class of items, we input the [CLS] representation into a fully connected layer as follows:
\begin{equation}
    p = \sigma(\mathbf{W}C + \mathbf{b})
\end{equation}
where $\mathbf{W} \in \mathbb{R}^{d \times n_c}$ is a weighed matrix in the shape of ${d \times n_c}$. $d$ is the word embedding dimension, also called hidden size, of BERT and $n_c$ is the number of classes in current task. $\mathbf{b}\in \mathbb{R}^{n_c}$ is a bias vector and $\sigma()$ is an activate function. 

With predicted $p$ and input label $l \in \mathbb{R} ^{n_c}$, we fine-tune the model with a cross-entropy loss. We assign an unique identity to each class beginning with $0$. For an item belonging to the $i$th class, $l[i] =1$ and others are $0$. 

\textbf{Base+PKGM model}
To enhance item classification with our PKGM, for each item, we provide service with $k$ presentations from both triple query module and relation query module, $2\times k$ in total. We first add a [SEP] symbol at the end of the title text and then input the $2\times k$ service vectors from PKGM into BERT, and we named it as \emph{Base$_{PKGM-all}$}. Embedding look up is unnecessary for service vectors and they are directly appended after the last token embedding of the input sentence.  Details are shown in Figure~\ref{fig:model-classification}.

We also fine-tune the Base+PKGM model with only $k$ triple query module representations to explore how much of a role the \textit{triple query service} plays in the model. The final $k$ title texts are replaced with service vectors from triple query module, and we abbreviate this model as \emph{Base$_{PKGM-T}$}. In the same way, \emph{Base$_{PKGM-R}$} refers to replacing final $k$ title texts with $k$ service vectors from relation query module.

\subsubsection{Dataset}
In this experiment, we take item categories as target classes. In our platform, each item is assigned to at least one category, and we maintain an item category tree to better organize billions of items and help recall items for a lot of business, like selecting items for the online market in a special event like Double Eleven. Thus we always try to assign at least one category for each item. 

Experiment dataset contains 1293 categories. Details are shown in Table~\ref{data:classification}. The number of positive samples and negative samples are $1:1$. To show the power of pre-trained models, both language model and knowledge graph model, with which we could get good performance on downstream tasks with a few training samples, we constrain the instance of each category less than 100 during data preparation.

\input{tables/item-classification-data}

\subsubsection{Experiment details}
We take the pre-trained BERT$_{BASE}$ model trained on Chinese simplified and traditional corpus\footnote{Released at https://storage.googleapis.com/bert\_models/2018\_11\_03/chinese\_L-12\_H-768\_A-12.zip}, whose number of layers is $12$, hidden size is $768$ and the number of attention heads is $12$. 

Following BERT, we add a special classification token [CLS] ahead of the input sequence, which is used as the aggregate sequence representation in the final hidden state. In this experiment, we set the length of the sequence of titles to 128. For titles shorter than $127$ (with [CLS] symbol excepted), we padding it with several zero embeddings to the required length as did in BERT, while for titles longer than $127$, we reserve the first $127$ words.

For BERT$_{PKGM-all}$ model, we replace the last $2\times k$ embeddings with service vectors from PKGM,  and for  BERT$_{PKGM-T}$ model and BERT$_{PKGM-R}$, we replace the last $k$ embeddings with service vectors from triple query module and relation query module respectively.

We fine-tune BERT with $3$ epochs with batch size as $32$ and learning rate as $2e^{-5}$. All parameters in BERT are unfix and representations from PKGM fixed during fine-tune.

\begin{figure*}[!htbp]
     \centering
     \subfigure[BERT.]{
         \centering
         \includegraphics[height=5.4cm]{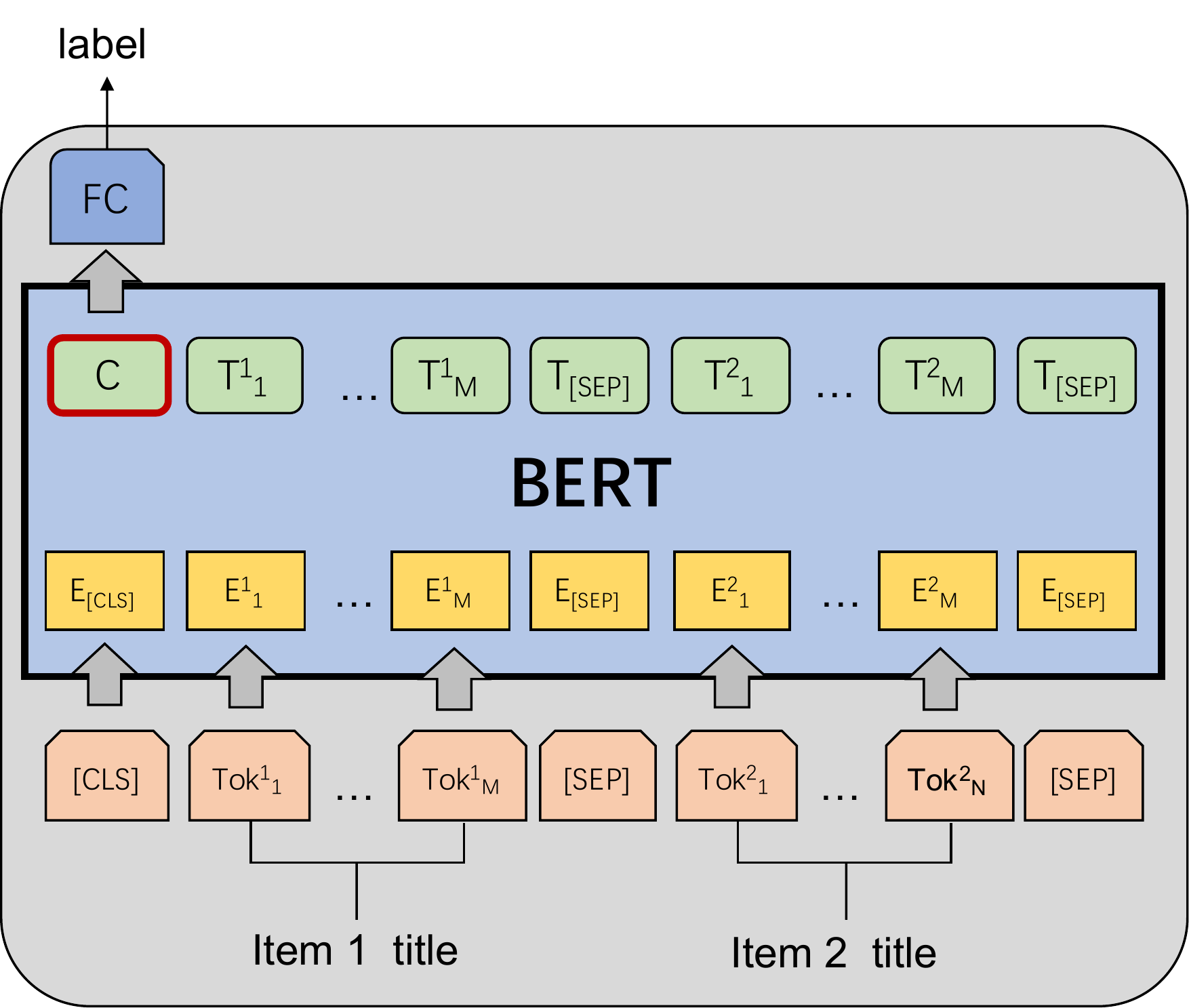}
         \label{fig:same-base}
     }
     \subfigure[BERT$_{PKGM}$.]{
         \centering
         \includegraphics[height=5.4cm]{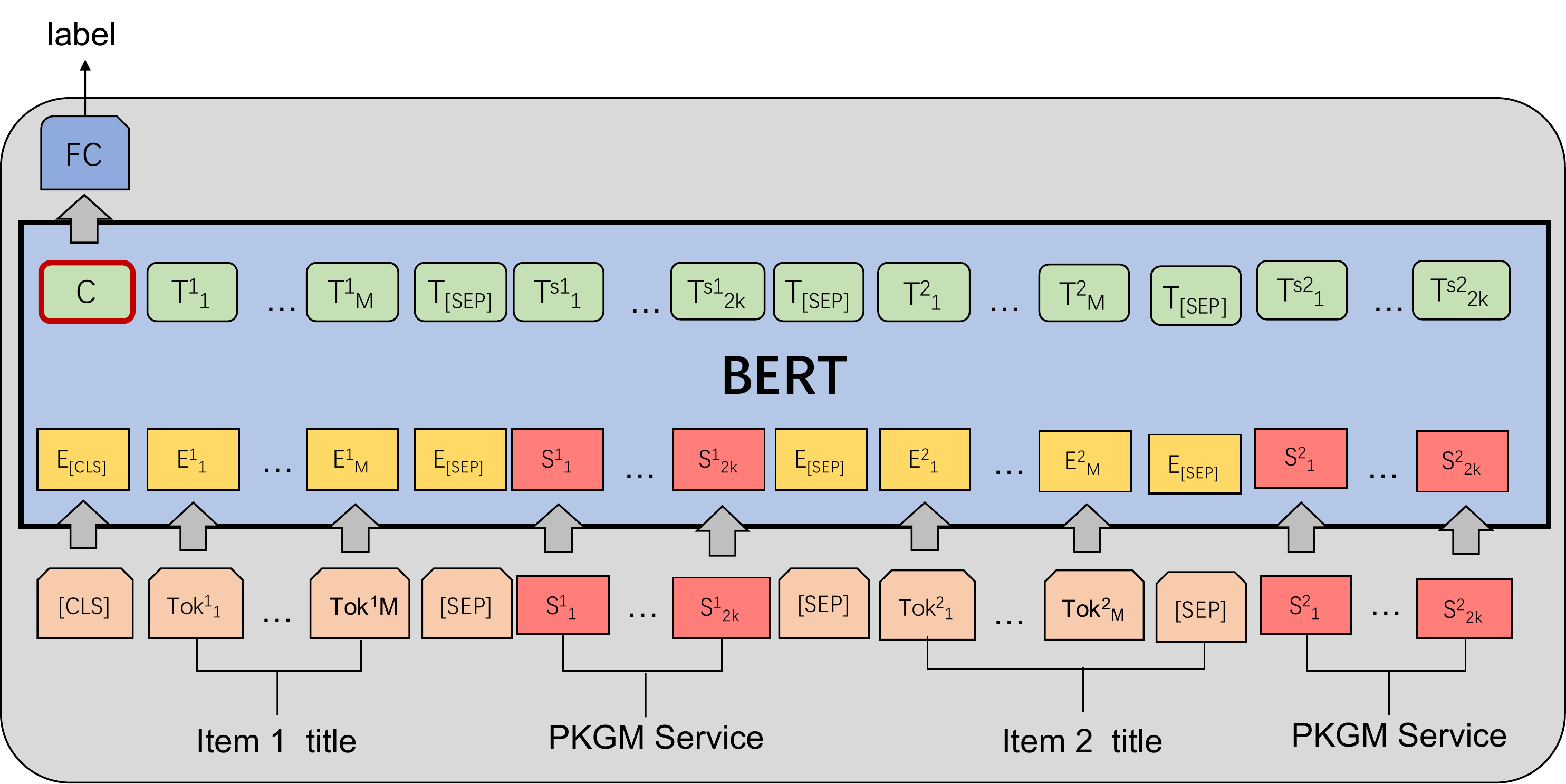}
         \label{fig:same-base+PKGM}
     }
    \caption{Models for item alignment.}
    \label{fig:model-same}
\end{figure*}

To evaluate performances, we report the accuracy of item classification in TABLE~\ref{res:classification}. We abbreviate the accuracy as AC and it is reported as the percentage from the validation dataset. Because the amount of categories is too large compared with other traditional classification tasks, we also report metrics $Hit@k(k=1,3,10)$ to describe the validity of the experimental results more accurately. We calculate $Hit@k$ by getting the rank of the correct label as its predicting category rank and $Hit@k$ is the percentage of test samples with prediction rank within $k$. 

\subsubsection{Results}
\input{tables/item-clasification-results}
The results are shown in Table~\ref{res:classification}, showing that BERT$_{PKGM}$ outperforms BERT on all of these three datasets, on both $Hit@k$ and prediction accuracy metrics. Specifically, BERT$_{PKGM-all}$ has the best performance of $Hit@1, Hit@3$ and $Hit@10$ and BERT$_{PKGM-R}$ achieves best accuracy. 
% We also observe that results of BERT$_{PKGM-all}$ and BERT$_{PKGM-R}$ perform greatly on $Hit@3$, $Hit@10$, and one of them always can reach the best experimental results. 
These results demonstrate the effectiveness of information from PKGM, among which relation query service vectors contributes more useful information than triple query service vectors.

Besides, BERT$_{PKGM-R}$ sometimes outperforms BERT$_{PKGM-all}$, which violates the idea that more features matter. We think it's probably because words in the original title play a more important role in the item classification task than service vectors provided from PKGM since less words are replaced in BERT$_{PKGM-R}$ than BERT$_{PKGM-all}$. 

%% file: tables/item-classification-data.tex
\begin{table}[!hbpt]
    \centering
    \caption{Data statistic for  item classification task}
    \begin{tabular}{c |c | c | c |c}
    \toprule
         & \# category & \# Train & \# Test & \# Dev \\
         \midrule
        dataset  &1293  & 169039 & 36225 & 36223  \\
        \bottomrule
    \end{tabular}
    \label{data:classification}
\end{table}

%% file: tables/item-clasification-results.tex
% \begin{table}[!hbpt]
%     \centering
%     \caption{Results for item classification task}
%     \begin{tabular}{l |l|c | c |c | c}
%     \toprule
%         Dataset & Method & Hit@1 &  Hit@3 & Hit@10 & AC  \\
%           \midrule
%         \multirow{2}{*}{dataset-100} & BERT & 71.03 & 84.91 & 92.47 & 71.52 \\
%         & BERT$_{PKGM-T}$ & 71.26 & 85.76 & 93.07 & 72.14\\
%         & BERT$_{PKGM-R}$ & 71.55 & 85.43 & 92.86 & \textbf{72.26}\\
%         & BERT$_{PKGM-all}$ & \textbf{71.64} & \textbf{85.90} & \textbf{93.17} & 72.19\\
%         \midrule
%         \multirow{2}{*}{dataset-50} & BERT & 60.98 & 78.99 & 89.21 & 59.06 \\
%         & BERT$_{PKGM-T}$ & 61.67 & 79.04 & 90.08 & 62.74 \\
%         & BERT$_{PKGM-R}$ & 61.52 & \textbf{80.09} & \textbf{90.39} & \textbf{62.98}\\
%         & BERT$_{PKGM-all}$ & \textbf{61.54} & 79.89 & 90.36 & 62.71\\
%         \midrule
%         \multirow{2}{*}{dataset-20} & BERT & 30.26 & 46.97 & 68.12 & 28.90 \\
%         & BERT$_{PKGM-T}$ & 30.65 & 47.80 & 67.40 & 29.62\\
%         & BERT$_{PKGM-R}$ & 31.47 & \textbf{50.69} & 69.07 & 30.63\\
%         & BERT$_{PKGM-all}$ & \textbf{32.09} & 50.19 & \textbf{70.07} & \textbf{30.91}\\
%         \bottomrule
%     \end{tabular}
%     \label{res:classification}
% \end{table}

\begin{table}[!hbpt]
    \centering
    \caption{Results for item classification task}
    \begin{tabular}{l|c | c |c | c}
    \toprule
         Method & Hit@1 &  Hit@3 & Hit@10 & AC  \\
          \midrule
          BERT & 71.03 & 84.91 & 92.47 & 71.52 \\
         BERT$_{PKGM-T}$ & 71.26 & 85.76 & 93.07 & 72.14\\
         BERT$_{PKGM-R}$ & 71.55 & 85.43 & 92.86 & \textbf{72.26}\\
         BERT$_{PKGM-all}$ & \textbf{71.64} & \textbf{85.90} & \textbf{93.17} & 72.19\\
        % \midrule
        % \multirow{2}{*}{dataset-50} & BERT & 60.98 & 78.99 & 89.21 & 59.06 \\
        % & BERT$_{PKGM-T}$ & 61.67 & 79.04 & 90.08 & 62.74 \\
        % & BERT$_{PKGM-R}$ & 61.52 & \textbf{80.09} & \textbf{90.39} & \textbf{62.98}\\
        % & BERT$_{PKGM-all}$ & \textbf{61.54} & 79.89 & 90.36 & 62.71\\
        % \midrule
        % \multirow{2}{*}{dataset-20} & BERT & 30.26 & 46.97 & 68.12 & 28.90 \\
        % & BERT$_{PKGM-T}$ & 30.65 & 47.80 & 67.40 & 29.62\\
        % & BERT$_{PKGM-R}$ & 31.47 & \textbf{50.69} & 69.07 & 30.63\\
        % & BERT$_{PKGM-all}$ & \textbf{32.09} & 50.19 & \textbf{70.07} & \textbf{30.91}\\
        \bottomrule
    \end{tabular}
    \label{res:classification}
\end{table}

%% file: C3-E-3Same_product_identification.tex
\subsection{Product Alignment}

\subsubsection{Task definition}
The target of the item alignment task is to find two items referring to the same product. For example, there are many online shops selling IPhoneXI with Green color and 256 GB capacity. They are different items on the platform while from the perspective of the product, they refer to the same one. Detecting items belonging to the same product contributes to daily business greatly, for example, recommending aligned items of items that a user is reviewing, which  help users deeply compare the prices, services after selling, and so on. More importantly, the number of items are much larger than the number of products, organizing items from the perspective of the product helps reduce the load of data management and mining.

Similar to item classification, item titles are used for alignment. And item alignment task could be framed as a paraphrase identification task, to distinguish whether two input sentences express the same meaning\cite{PI}. In the context of the item alignment, the target is to distinguish whether two input item titles describe the same product. The prevailing approach of training and evaluating paraphrase identification models are constructed as a binary classification problem: the model is given a pair of sentences and is judged by how accurately it classifies pairs as either paraphrases or non-paraphrases. Thus a mapping function $f: \mathcal{P_T} \mapsto \mathcal{C}$ will be learned, in which $\mathcal{P_T}$ is a set of item title pairs and   $\mathcal{T}$ is a set of item title. $\mathcal{C}$ contains binary classes $True$ and $False$. Similar to item classification, for each item title $t\in \mathcal{T}$, $t = [ w_1, w_2, w_3, ..., w_n ]$ is an ordered sequence composed of words.

\subsubsection{Model}

\textbf{Base model}
Paraphrase identification is a well-studied sentence pair modeling task and is very useful for many NLP applications such as machine translation(MT)\cite{vaswani2018tensor2tensor}, question answering(QA)\cite{yih2015semantic} and information retrieval(IR)\cite{berger2017information}. Many methods have been proposed for it in recent years including pairwise word interaction modeling with deep neural network system~\cite{DBLP:conf/naacl/HeL16}, character level neural network model~\cite{DBLP:conf/naacl/LanX18}, and pre-trained language model~\cite{BERT}. Since the pre-trained language models have reached state-of-the-art performance on paraphrase identification tasks, we apply BERT as the base model for the same alignment. 

We show the details of applying BERT on the item alignment in Figure~\ref{fig:model-same}. The title of two items are input into BERT as a sentence pair and we take the representation corresponding to [CLS] symbol as the representation of the sentence pair which will be used for binary classification with a fully connected layer, as done in item classification task. 

\textbf{Base+PKGM}
For each item, we provide service with $2\times k$ vectors from the triple query module and relation query module of PKGM to enhance the item alignment task with item knowledge. Similar to the operation in item classification, we add a [SEP] symbol at the end of each title text and  $4\times k$ service vectors are added to BERT finally. After that, we concatenate two-sentence input together. Details are shown in Figure~\ref{fig:model-same}. Similar to item classification, with all service vectors applied, we call the model as \emph{Base$_{PKGM-all}$}. \emph{Base$_{PKGM-T}$} and \emph{Base$_{PKGM-R}$} are defined in the same way which refer to input the $k$ service vectors from triple query module and relation query module in PKGM respectively. In this task, \emph{Base} refers to BERT.

\subsubsection{Dataset}
We experiment on $3$ datasets containing different types of items. In our platform, item alignment is only necessary to distinguish items with the same type since items with different types refer to different products for sure. The first dataset is generated from skirts for girls, the second one from hair decorations and the last one from children's socks.  In each dataset, one data sample includes two item titles and a label, in which samples belonging to the same products are labeled with $1$ and $0$ otherwise. There are less than 10 thousand training samples in each dataset and we keep the number of item pairs in train/test/dev dataset as $7:1.5:1.5$. Details of these datasets are shown in Table~\ref{tab:data-same-item}.

\input{tables/same-item-detection-data}

\subsubsection{Experiment details}
Consistent with the previous item classification task, we take the same pre-trained BERT$_{BASE}$ model and the same input format, except for some difference in the input data. The input embedding sequence consists of two different items. A [CLS] symbol is added in the beginning of the title sequence  and one [SEP] symbol is appended at the last position of each title sequence to differential these two items. Since the whole sequence length is 128, the length of each item sequence is restricted within 63 and we adopt the same method to formalize the item sequence with item classification task.

% evaluation metrics
We report prediction accuracy and metrics $Hit@k(k=1,3,10)$ to evaluate the performance on item alignment. For metric $Hit@k$, we generate $n$ negative triples by replacing one of the aligned items randomly and together with the original aligned pairs, we will get $n+1$ prediction results. Afterward, we rank $n+1$ prediction probabilities in ascend order and get the rank of the aligned pairs as its prediction rank. $Hit@k$ is the percentage of test samples with prediction rank within $k$. In this case, the number of negative triples for each aligned pair is 99 and in other words, we get prediction rank from 100 candidate samples.

\subsubsection{Results}

Table~\ref{tab:res-same-item-R} shows the rank results for item alignment. BERT$_{PKGM-all}$ model outperforms BERT model on $Hit@10$ metrics on all three datasets and it has the best performance at both category-2 and category-3 with all $Hit@k$ metrics. These results show the effectiveness of PKGM and it promotes the accuracy of the prediction better. In the meantime, BERT model has a weak advantage over BERT$_{PKGM-all}$ model at the category-1 on $Hit@1$ metrics and this could probably be due to the larger amount of data in this category. To a certain extent, the title text with enough training examples could make the model learn item information better, while the PKGM could play a greater role in the case of a small amount of data relatively.

Table~\ref{tab:res-same-item-C} shows the accuracy results. Obviously, BERT$_{PKGM-all}$ has the best performance on all datasets and it convincingly demonstrates the PKGM could promote the effect on item alignment task.

\input{tables/same-item-detection-results}

%% file: tables/same-item-detection-data.tex
% \begin{table}[!hbpt]
%     \centering
%     \caption{\ye{(modified by @ye)Data statistic for same item detection task}}
%     \begin{tabular}{l |c | c | c }
%     \toprule
%           & \# Train & \# Test & \# Dev \\
%         %  \midrule
%         % dataset  &1293  & 169039 & 36225 & 36223  \\
%         \midrule
%         % cate 0 & 18923 &4057  & 4055 \\  
%         % cate 1 & 10687 & 2292 & 2290 \\
%         % cate 2 & 10156 &2178 & 2177\\
%         % cate 3 & 5284 & 1134 & 1133 \\
%         4 - skirts for girls & 9463  & 2030 & 2028  \\
%         5 - hair decorations & 4850 & 1042 & 1040  \\
%         % cate 6 & 8394 & 1800 & 1799 \\
%         % cate 7 & 6560 & 1407 & 1406 \\
%         % cate 8 & 8178 & 1754 & 1753 \\
%         9 - Children's socks & 3969 & 853 & 851  \\
%         \bottomrule
%     \end{tabular}
%     \label{tab:data-same-item}
% \end{table}

\begin{table}[!hbpt]
    \centering
    \caption{Data statistic for item alignment task.}
    \begin{tabular}{l |c | c | c | c | c}
    \toprule
          & \# Train & \# Test-C & \# Dev-C & \# Test-R & \# Dev-R\\
        %  \midrule
        % dataset  &1293  & 169039 & 36225 & 36223  \\
        \midrule
        % cate 0 & 18923 &4057  & 4055 \\  
        % cate 1 & 10687 & 2292 & 2290 \\
        % cate 2 & 10156 &2178 & 2177\\
        % cate 3 & 5284 & 1134 & 1133 \\
        category-1 & 4731 & 1014 & 1013 & 513 & 497\\
        category-2 & 2424 & 520 & 519 & 268 & 278 \\
        % cate 6 & 8394 & 1800 & 1799 \\
        % cate 7 & 6560 & 1407 & 1406 \\
        % cate 8 & 8178 & 1754 & 1753 \\
        category-3 & 3968 & 852 & 850 & 417 & 440\\
        \bottomrule
    \end{tabular}
    \label{tab:data-same-item}
\end{table}

%% file: tables/same-item-detection-results.tex
\begin{table}[!hbpt]
    \centering
    \caption{Hit@$k$ results for item alignment.}
    \begin{tabular}{l |l|c | c|c  }
    \toprule
         Method & dataset  & Hit@1 &Hit@3 & Hit@10 \\
        \midrule
        BERT  & \multirow{2}{*}{category-1} & \textbf{65.06} & 76.06 & 86.68  \\
        BERT$_{PKGM-all}$ & & 64.75 &\textbf{77.50} & \textbf{87.43}  \\
        \midrule
        BERT  & \multirow{2}{*}{category-2} & 65.86 & 78.07 & 87.59  \\
        BERT$_{PKGM-all}$ & & \textbf{66.13} & \textbf{78.19} & \textbf{87.96}  \\
        % \midrule
        % BERT  & \multirow{2}{*}{cate 9 - 1985 - half} & 50.95 & 65.94 & 81.05 & 85.05 \\
        % BERT$_{PKG}$ & & 50.60 & 64.74 & 79.97 & 86.00 \\
        \midrule
        BERT  & \multirow{2}{*}{category-3} & 49.64 & 66.18 & 82.37 \\
        BERT$_{PKGM-all}$ & & \textbf{50.60} & \textbf{67.14} & \textbf{83.45} \\
%         BERT$_{PKG}$ &  & & &  & & & 50.60 & 67.14 & 83.45 &87.88\\
        \bottomrule
    \end{tabular}
    \label{tab:res-same-item-R}
\end{table}

\begin{table}[]
    \centering
    \caption{Accuracy results for item alignment.}
    \begin{tabular}{l|c c c }
        \toprule
          & category-1 & category-2 & category-3 \\
        \midrule
         BERT& 88.94 & 89.31  & 86.94 \\
         BERT$_{PKGM-T}$ & 88.65 & 89.89 & 87.88 \\
         BERT$_{PKGM-R}$ & 89.09 & 89.60 & 87.88 \\
         BERT$_{PKGM-all}$& \textbf{89.15} & \textbf{90.08} & \textbf{88.13} \\
        \bottomrule
    \end{tabular}
    \label{tab:res-same-item-C}
\end{table}

%% file: C3-E-4Item_recommendation.tex
\subsection{Item Recommendation}
\begin{figure*}[!htbp]
     \centering
     \subfigure[NCF.]{
         \centering
         \includegraphics[height=6.7cm]{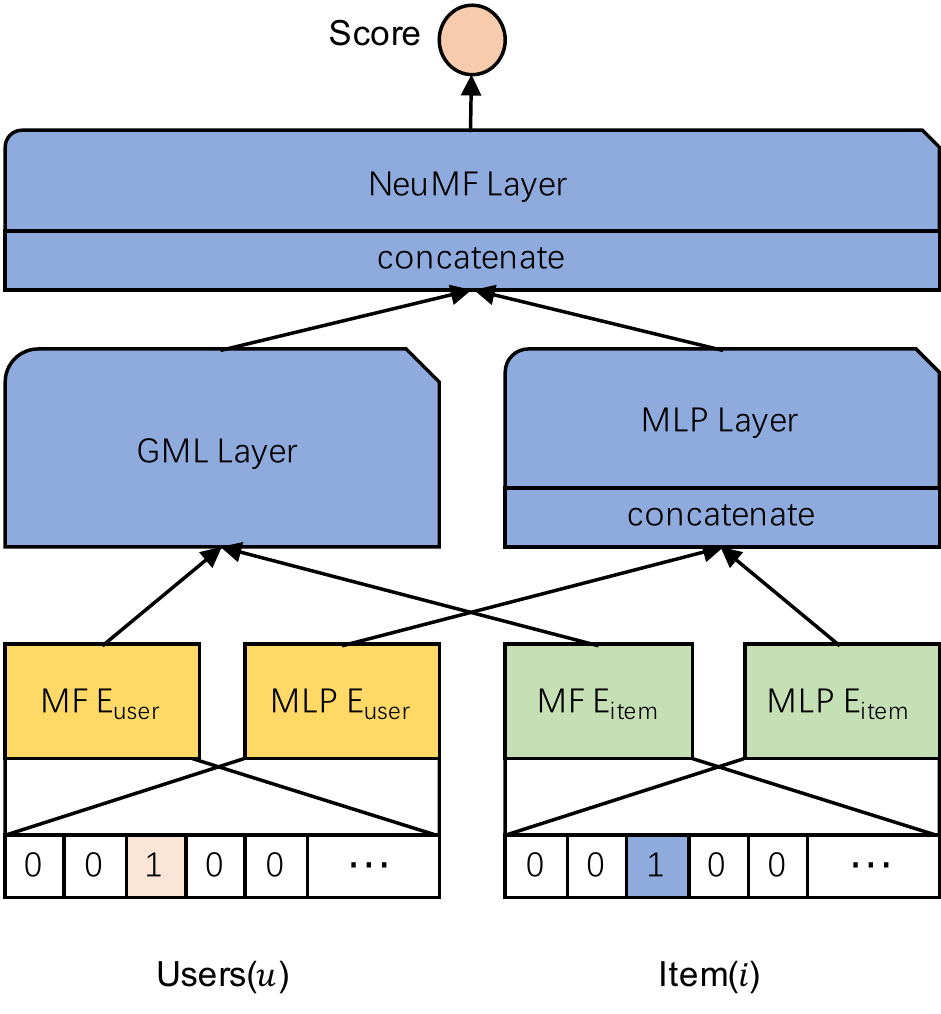}
         \label{fig:recommendation-base}
     }
     \subfigure[NCF$_{PKGM}$.]{
         \centering
         \includegraphics[height=6.7cm]{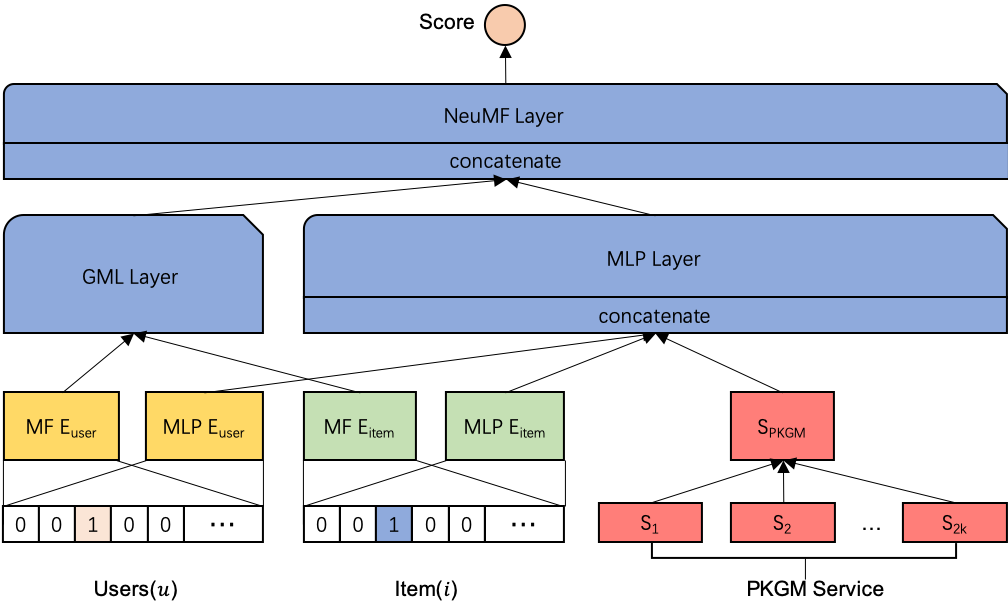}
         \label{fig:recommendation-base+PKGM}
     }
    \caption{Models for item recommendation.}
    \label{fig:model-recommendation}
\end{figure*}
\subsubsection{Task definition}
Item recommendation aims to properly recommend items to users that users will interact with a high probability. As implicit feedback (click, buy, etc) is more commonly observed in modern e-commerce platforms, we consider only implicit feedback for recommending items to corresponding users with the help of service vectors provided by PKGM. To learn user preference and make proper recommendations from observed implicit user feedback, we treat it as a ranking problem like described in \cite{DBLP:conf/www/HeLZNHC17}.

\subsubsection{Model}

\textbf{Base model}
We make use of NCF\cite{DBLP:conf/www/HeLZNHC17} as a general framework of our base model. In NCF, GMF layer and MLP layer are used for modeling item-user interactions, in which GMF uses a linear kernel to model the latent feature interactions, and MLP uses a non-linear kernel to learn the interaction function from data. There are 4 layers in NCF including input layer, embedding layer, neural CF layer, and output layer. 

The input layer contains user and item sparse feature vectors $\mathbf{v}_u^U$ and $\mathbf{v}_i^I$ that describe user $u$ and item $i$ respectively. Since we conduct the downstream recommendation experiment based on a pure collaborative setting, we make $\mathbf{v}_u^U$ and $\mathbf{v}_i^I$ one-hot sparse vector generated according to the identity of  $u$ and $i$ in the dataset.

Above the input layer, there is an embedding layer which is a fully connected layer to project the sparse input representation to a dense vector, called embedding.  User embedding and item embedding could be regarded as the latent vector for users and items in the context of latent factor models. 
\begin{equation}
    \mathbf{p}_u = \mathbf{P}^T\mathbf{v}_u^{U}, \; \mathbf{q}_i = \mathbf{Q}^T\mathbf{v}_i^I
\end{equation}

Then user embedding $\mathbf{p}_u$ and item embedding $\mathbf{q}_i$ are fed into a neural collaborative filtering layer, to map the latent vectors to prediction scores. Thus the NCF predictive model could be formulated as 
\begin{equation}
\hat{y}_{ui} = f_{NCF}(\mathbf{p}_u, \mathbf{q}_i| \mathbf{P}, \mathbf{Q}, \Theta_f),
\end{equation}
where $\mathbf{P} \in \mathbb{R}^{M \times K}$ and $\mathbf{Q} \in \mathbb{R}^{N \times K}$ are embedding matrix for users and items respectively. And $\Theta_f$ is a set of parameters in function $f_{NCF}$. $f_{NCF}$ is a user-item interaction function that makes $\mathbf{p}_u$ and $\mathbf{q}_i$ interact and influence with each other and finally gives a score for the input item-user pair. And there are two ways to make such interactions, one is matrix factorization (MF) and the other one is multi-layer perceptron (MLP). 

MF is the most popular model for recommendation and has been investigated extensively in literature\cite{MF}. The mapping function in  neural generalized matrix factorization (GMF) layer is 
\begin{equation}
    \phi^{GMF}(\mathbf{p}_u, \mathbf{q}_i) = \mathbf{p}_u \circ \mathbf{q}_i ,
\end{equation}
where $\circ$ denotes the element-wise product of vectors. 

Using two pathways to model users and items has been widely adopted in multimodal deep learning models~\cite{RS1,RS2}. Thus in the MLP interaction layer, we first make a vector concatenation of $\mathbf{p}_u$ and $\mathbf{q}_i$
\begin{equation}
    \mathbf{z}_1 = \phi_1^{MLP}(\mathbf{p}_u, \mathbf{q}_i) = {
\left[ \begin{array}{ccc}
\mathbf{p}_u \\
\mathbf{q}_i
\end{array} 
\right ]}
\end{equation}
then we add hidden layers on the concatenated vector, using a standard MLP to learn the interaction between user and item latent features. This endows the model a large level of flexibility and non-linearity to learn the interactions between $\mathbf{p}_u$ and $\mathbf{q}_i$
% \begin{equation}
\begin{align}
    \phi_2^{MLP}(\mathbf{z}_1) & = a_2 (\mathbf{W}_2 ^T \mathbf{z}_1 + \mathbf{b}_2) \\
    & ...... \\
    \phi_L^{MLP}(\mathbf{z}_{L-1}) & = a_{L} (\mathbf{W}_L ^T \mathbf{z}_{L-1} + \mathbf{b}_L) 
\end{align}
% \end{equation}
where $\mathbf{W}_x$, $\mathbf{b}_x$ and $a_x$ denote the weight matrix, bias vector, and activation function for the $x$-th layer's perceptron, respectively.

\input{tables/recommendation-results}

So far we have two interaction vectors for the input user and item embeddings, then we fuse these two vectors as follows:
\begin{equation}
\hat{y}_{ui} = \sigma(\mathbf{h}^T {
\left[ \begin{array}{ccc}
\phi^{GMF} \\
\phi^{MLP} 
\end{array} 
\right ]})
\end{equation}
The objective function to minimize in NCF method is as follows
\begin{equation}\label{equation-19}
    L = - \sum_{(u,i) \in \mathcal{Y}} y_{ui} log(\hat{y}_{ui}) + (1-y_{ui})log(1-\hat{y}_{ui})
\end{equation}
where $y_{ui}$ is the label for input user $u$ and item $i$, and if they have interaction in the dataset $y_{ui}=1$ and $y_{ui}=0$ otherwise. $\mathcal{Y}$ is the training set containing positive pairs and negative pairs. Since the original dataset only contains positive pairs, we take the same strategy used in \cite{DBLP:conf/www/HeLZNHC17} to generate negative pairs. Negative pairs are uniformly sampled from unobserved interactions for each positive pair, and replace the positive item with negative items that are unobserved. We refer negative sampling ratio as the number of negative pairs for each positive pair.

\textbf{Base+PKGM}
Since NCF is built based on item embeddings, we integrate  vector services from PKGM with item embeddings in the second way as introduced before. Specifically, for each user-item pair, PKGM with provide $2\times k$ vectors for the item, denoted as $[S_1, S_2, ..., S_{2k}]$. We first combine these vectors into one as follows
\begin{equation}
    S_{PKGM} = \frac{1}{k} \sum_{i \in [1, 2, ..., k]} [S_i;S_{i + k}]
\end{equation}
then we integrate $S_{PKGM}$ into MLP layer by making $\mathbf{z}_1$ as a concatenation results of three vectors
\begin{equation}
    \mathbf{z}_1 = \phi _1 ^{MLP} (\mathbf{p}_u, \mathbf{q}_i, S_{PKGM}) = {
\left[ \begin{array}{ccc}
\mathbf{p}_u \\
\mathbf{q}_i \\
S_{PKGM} 
\end{array} 
\right ]}.
\end{equation}
Other parts of NCF stay the same. Fig.~\ref{fig:model-recommendation} shows the general framework of NCF and NCF$_{PKGM}$. 

\subsubsection{Dataset}
We conduct experiments on real data sampled from records on Taobao platform. The statistics of our sample data are shown in TABLE \ref{tab:taobao-recommendation-statistics}. There are 29015 users and 37847 items with 443425 interactions between them. The number of interactions for each user is assured of being more than 10. Each interaction record is a user-item pair representing that the user has an interaction with the item.
\input{tables/recommendation-data}

\subsubsection{Experiment details}
% evaluation protocols and so on 
The performance of item recommendation is evaluated by the \textit{leave-one-out} strategy which is used in \cite{DBLP:conf/www/HeLZNHC17}. For each user, we hold-out the latest interaction as the test set and others are used in the train set. As for the testing procedure, we uniformly sample 100 unobserved negative items, ranking the positive test item with negative ones together. In this way, $Hit\,Ratio$(HR)@k and $Normalized\ Discounted\ Cumulative\ Gain $(NDCG)@k are used as the final metrics of ranking where $k \in \{1,3,5,10,30\}$. Both metrics are computed for each test user, and average scores are regarded as final results.

% parameter settings
We randomly sample one interaction for each user and it is treated as a validation sample to determine the best hyper-parameters of the model. For the GMF layer part, the dimension of user embedding and item embedding are both set to $8$. As for MLP layers, the dimension of user embedding and item embedding are both set to $32$. Three hidden layers of size $[32, 16, 8]$ after embedding layers are used for both NCF and NCF$_{PKGM}$. For NCF$_{PKGM}$, there are external PKGM features fed to concatenate with MLP user embedding and MLP item embedding. The model is optimized with loss defined in Equation~\eqref{equation-19} with an external $L2$ regularization on user and item embedding layer in MLP and GMF. Regularization factor $\lambda$ is chosen as 0.001. Mini-batch Adam\cite{adam} is used for model training with a batch size of $256$ and the total number of epochs is $100$. Learning rate is set as $0.0001$. And the dimension of the prediction layer is $16$, which consists of two partial outputs of $8$ and $8$ from GMF layer and MLP layer respectively. In this work, we use a negative sampling ratio of $4$. For simplicity and effectiveness, we use the non-pre-train version of NCF model for base model and NCF$_{PKGM}$ as PKGM featured model.
\subsubsection{Results}

Experiment results are shown in TABLE~\ref{tab:res-recommendation}. 
% NCF with PKGM-* subscript indicates whether the augmented feature is produced by PKGM-triple query module only, PKGM-relation query module only, or both module. 
From the result we can see that:

Firstly, all of the PKGM feature augmented models outperform the NCF baseline in all metrics. For NCF$_{PKGM-T}$ model, it outperforms the NCF baseline with an average promotion of $0.37\%$ on Hit Ratio metrics and $0.0023$ on NDCG metrics. For NCF$_{PKGM-R}$ model, it outperforms the NCF baseline with an average promotion of $3.66\%$ on Hit Ratio metrics and $0.0343$ on NDCG metrics.
For NCF$_{PKGM-all}$ model, it outperforms the NCF baseline with an average promotion of $3.47\%$ on Hit Ratio metrics and $0.0324$ on NDCG metrics. This promotion illustrates that features learned by our PKGM model provide external information that are not included in original user-item interaction data and it is useful for downstream tasks like e-commerce recommendation.

Secondly, the key information of different service vectors provided by PKGM model are different. Performances on NCF$_{PKGM-R}$ are better than those on NCF$_{PKGM-T}$. Thus the information from relation query module is relatively more useful than the information from PKGM-triple query module in item recommendation scenario, which is largely due to the fact that properties are more effective than entities and values when modeling user-item interaction.

%% file: tables/recommendation-results.tex
\begin{table*}[!hbpt]
    \centering
    \caption{Results for item recommendation.}
    \begin{tabular}{l|c|c|c|c|c|c|c|c|c|c}
    \toprule
         & HR@1 & HR@3 & HR@5 & HR@10 & HR@30 & NDCG@1 & NDCG@3 & NDCG@5 &NDCG@10 & NDCG@30\\
        \midrule
        NCF  & 27.94 & 44.26 & 52.16 & 62.88 & 81.26 & 0.2794 & 0.3744 & 0.4069 & 0.4415 & 0.4853 \\
        \midrule
        NCF$_{PKGM-T}$ & 27.96 & 44.83 & 52.43 & 63.51 & 81.62 & 0.2796 & 0.3778 & 0.4091 & 0.4449 & 0.4880\\
        NCF$_{PKGM-R}$ & \textbf{31.01} & \textbf{47.99} & \textbf{56.10} & \textbf{66.98} & \textbf{84.73} & \textbf{0.3101} & \textbf{0.4091} & \textbf{0.4424} & \textbf{0.4777} & \textbf{0.5200}\\
        NCF$_{PKGM-all}$ & 30.76 & 47.92 & 55.60 & 66.84 & 84.71 & 0.3076 & 0.4079 & 0.4395 & 0.4758 & 0.5185 \\
        \bottomrule
    \end{tabular}
    \label{tab:res-recommendation}
\end{table*}

%% file: tables/recommendation-data.tex
\begin{table}[!hbpt]
    \centering
    \caption{Data statistics for item recommendation.}
    \begin{tabular}{c |c|c | c }
    \toprule
         & \# Items & \# Users & \# Interactions \\
         \midrule
        TAOBAO-Recommendation  & 37847 & 29015 & 443425 \\
        % \midrule
        % & & & \\
    \bottomrule
    \end{tabular}
    \label{tab:taobao-recommendation-statistics}
\end{table}

%% file: C4-Related_work.tex
\section{Related work}

\subsection{Knowledge Graph Embedding}
With the introduction of embedding entities and relations of the knowledge graph into continuous vector space by TransE~\cite{TransE-Bordes-2013}, the method of Knowledge Graph Embedding (KGE) has made great progress in knowledge graph completion and application. The key idea that transforming entities and relations in triple (\textit{h, r, t}) into continuous vector space aims to simplify the operation in KG. TransH \cite{TransH-Wang-2014} proposes that an entity has different representations under different relationships to overcome the limitations within multi-relation problems. TransR \cite{TransR-Lin-2015} introduces relation-specific spaces other than entity embedding spaces. TransD \cite{TransD-Ji-2015} decomposes the projection matrix into a product of two vectors. TranSparse \cite{TranSparse-Ji-2016} simplifies TransR by enforcing sparseness on the projection matrix. MuRP\cite{DBLP:conf/nips/BalazevicAH19} and ATTH\cite{DBLP:conf/acl/ChamiWJSRR20} lead hyperbolic geometry into embedding learning of knowledge graph to capture logical hierarchies and patterns. All of these models try to achieve better effects on multi-relation problems.

Different from the translational distance models mentioned above, the semantic matching model uses a similarity based scoring function. They measure the credibility of facts by matching the latent semantics of entities and the relationships, methods include RESCAL \cite{RESCAL-Nickel-2011}, DistMult \cite{DistMult-Yang-2015}, ComplEx \cite{ComplEx-Trouillon-2016} and ConvE \cite{ConvE-Dettmers-2018}, etc.

The knowledge graph embedding method learns representations for entities and relations that preserve the information contained in the graph. The method has been applied to various tasks and applications and achieved good results in the field of the knowledge graph, including link prediction\cite{CrossE}, rule learning\cite{IterE}, relation extraction\cite{LFDS}, and entity alignment\cite{sun2018bootstrapping}. We adopt the representation learning method in the  billion-scale-commerce product knowledge graph to calculate and to evaluate the relationship among items.

\subsection{Pre-train Language Models.}
The pre-trained language models have achieved excellent results in many NLP tasks and bring convenience in completing various specific downstream tasks, such as BERT \cite{BERT}, XLNet\cite{XLNet-Yang-2019}, and GPT-3\cite{gpt3_brown2020language}. Recently, more and more researchers pay attention to the combination of knowledge graph and pre-trained language model, to enable PLMs to reach better performance with the awareness of knowledge information. 

K-BERT \cite{K_BERT-Liu-2019} injects knowledge triples into a sentence to generate unified knowledge-rich language representation. 
ERNIE \cite{ERNIE-Zhang-2019} integrates entity representations from the knowledge module into the semantic module to represent heterogeneous information of tokens and entities into a united feature space. 
KEPLER \cite{KEPLER-Wang-2019} encodes textual descriptions for entities as text embeddings and treats the description embeddings as entity embeddings.
KnowBert \cite{KnowBert-Peters-2019} uses an integrated entity linker to generate knowledge enhanced entity-span representations via a form of word-to-entity attention.
K-Adapter \cite{wang2020K_Adapter} takes RoBERTa as the pre-trained model and injects factual knowledge and linguistic knowledge with a neural adapter for each kind of infused knowledge. 
DKPLM \cite{su2020DKPLM} could dynamically select and embed knowledge context according to the textual context for PLMs, with the awareness of both global and local KG information.
JAKET \cite{yu2020JAKET} proposes a joint pre-training framework, including the knowledge module to produce embeddings for entities in text and language modules to generate context-aware embeddings in the graph. What's more, KALM \cite{rosset2020KALM}, ProQA \cite{xiong2020progressively}, LIBERT\cite{lauscher2019informing} and other researchers explore the fusion experiment with knowledge graph and pre-trained language module in different application tasks.

Based on BERT or other pre-trained language models, these methods aim at adjusting the language model by injecting knowledge bases or completing knowledge graphs from textual representation. Whereas, we pre-train our model on the structural and topological information of a knowledge graph without language model and apply the model into several downstream tasks in e-commerce knowledge.

%% file: C5-Conclusion.tex
\section{Conclusion and future work}
In this work, we present our experience of pre-training knowledge graph to provide knowledge services for other tasks on the billion-scale e-commerce product knowledge graph. We propose Pre-trained Knowledge Graph Mode (PKGM) with two modules to handle triple queries and relation queries in vector space, which could provide  item knowledge services in a uniform way for embedding-based models without accessing triple data in KG. We also propose general ways of applying service vectors in different downstream task models. With three types of downstream tasks, we prove that PKGM successfully improves their performance. In future work, we would like to apply PKGM to more downstream tasks and explore other candidate ways to apply service vectors. 

%% file: C6-Acknowledgements.tex
\section{Acknowledgements}
This work is funded by NSFCU19B2027/91846204/61473260, national key research program 2018YFB1402800 and supported by Alibaba Group through Alibaba Innovative Research Program.